# RAVE CHECKLIST: RECOMMENDATIONS FOR OVERCOMING CHALLENGES IN RETROSPECTIVE SAFETY STUDIES OF AUTOMATED DRIVING SYSTEMS

John M. Scanlon[a,*], Eric R. Teoh[b], David G. Kidd[b], Kristofer D. Kusano[a], Jonas Bärgman[c], Geoffrey Chi-Johnston[d], Luigi Di Lillo[e,f,g], Francesca Favaro[a], Carol Flannagan[h], Henrik Liers[i], Bonnie Lin[d], Magdalena Lindman[j], Shane McLaughlin[k], Miguel Perez[l], Trent Victor[a]

[a]Waymo LLC, 1600 Amphitheatre Parkway, Mountain View, CA, USA

[b]Insurance Institute for Highway Safety, Arlington, VA, USA

[C]Chalmers University of Technology, Sweden

[d]Cruise LLC, CA, USA

[e]Swiss Reinsurance Company, Ltd, Reinsurance Solutions, Switzerland

[f]Autonomous Systems Laboratory, Stanford University, CA, USA

[g]Zardini Lab, Massachusetts Institute of Technology (MIT), MA, USA

[h]University of Michigan Transportation Research Institute (UMTRI), MI, USA,

[i]Verkehrsunfallforschung an der TU Dresden GmbH (VUFO), Germany

[j]If P&C Insurance, Sweden

[k]Torc Robotics, Blacksburg, VA, USA

[l]Virginia Tech Transportation Institute (VTTI), Blacksburg, VA, USA

*Preprint Version - The paper has been submitted to journal.*

**ABSTRACT**

**Objective**

The public, regulators, and domain experts alike seek to understand the effect of deployed SAE level 4 automated driving system (ADS) technologies on safety. The recent expansion of ADS technology deployments is paving the way for early stage safety impact evaluations, whereby the observational data from both an ADS and a representative benchmark fleet are compared to quantify safety performance.

**Methods**

In January 2024, a working group of experts across academia, insurance, and industry came together in Washington, DC to discuss the current and future challenges in performing such evaluations. A subset of this working group then met, virtually, on multiple occasions to produce this paper.

**Results**

This paper presents the RAVE (Retrospective Automated Vehicle Evaluation) checklist, a set of fifteen recommendations for performing and evaluating retrospective ADS performance comparisons. The recommendations are centered around the concepts of (1) quality and validity, (2) transparency, and (3) interpretation.

**Conclusion**

Over time, it is anticipated there will be a large and varied body of work evaluating the observed performance of these ADS fleets. Establishing and promoting good scientific practices benefits the work of stakeholders, many of whom may not be subject matter experts. This working group's intentions are to: i) strengthen individual research studies and ii) make the at-large community more informed on how to evaluate this collective body of work.

**Keywords**: Automated Driving Systems, Safety Impact Analysis, ADS benchmarking, Traffic Safety

## INTRODUCTION

Retrospectively evaluating the collision prevention and mitigation safety impact of an automated driving system (ADS) deployment is a relatively new area for transportation researchers. To address the need for retrospective methods in this space, some studies leverage decades of traffic safety methods and literature while others build on methods from other domains. However, there are no established best practices in this area, which makes the interpretation of results difficult for parties with limited expertise in retrospective methods and/or ADS safety impact evaluations. Yet, as ADS deployments continue to expand, there is value in examining the methodological process and providing all stakeholders with an organized set of simple, clear recommendations for conducting and evaluating these safety impact assessments.

A working group of traffic safety experts, striving for consensus on methodological choices and interpretations of retrospective ADS safety assessment studies, was formed and set out to identify and discuss the ongoing challenges facing the community. The working group was also motivated by the long history of epidemiological work using observational data outside of traffic safety, where a simple set of recommended practices could benefit the community at-large (Von Elm et al., 2014). This publication is in no way intended to be prescriptive as to what data and methodology should be relied upon, but instead identifies and recommends best practices for subsequent retrospective studies. We, as researchers, recognize the need for novel methodological approaches and also acknowledge the inherent limitations that will undoubtedly accompany *any* safety impact study that is put forward. Nevertheless, all studies are not equivalent in their merits, and stakeholders need to be vigilant around retrospective safety assessment in order to better execute research and correctly interpret the results. This research, notably, enables informed policy decisions, so all stakeholders should have a shared interest in understanding the safety performance of ADS based on the strongest possible evidence.

### The Working Group

Experts from across industry, insurance, and academia formed a working group on the topic of ADS retrospective safety impact. This included a two-day meeting in January 2024 in Washington, DC with a goal to broadly identify and align opinions on the foreseeable challenges in conducting and interpreting the results that the community faces when performing retrospective safety impact assessments. In advance, participants were encouraged to prepare for discussion on important methods, technical challenges, and recommendations for conducting retrospective safety impact analyses for ADS. During this meeting, members of the working group shared their individual perspectives, discussed a wide range of technical considerations, and began to develop recommendations. A subset of the working group then met, virtually, on multiple occasions to iterate on content and draft this document. The result is a list of recommendations and explanations for performing or evaluating a retrospective ADS safety impact analysis that is based on the collective expertise of the working group.

### Clarification of Terms and Application

This paper focuses on recommendations for the evaluation of SAE level 4 systems, which are ADS-equipped vehicles that can handle the entire dynamic driving task without a human behind the wheel (SAE, 2021). In this paper, as a shorthand, these SAE level 4 systems are just referred to as “ADS”, which formally encompasses SAE level 3 and 5 systems as well. Today, SAE level 4 systems are being operated as part of fleets, meaning the vehicles and the ADS are maintained and operated by companies rather than individual consumers. A portion of the recommendations herein will likely apply to evaluations of future SAE level 5 systems and maybe applicable for ADS level 3 system evaluations. However, evaluating SAE level 3 systems may introduce other challenges related to having a human-in-the-loop that are beyond the scope of this paper, such as relying on consumers to report crashes, intermittent use of the automation with transitions between automated and manual driving, and even the mere presence of a human driver during automated driving that has the option to take over control (Schwall et al., 2020).

SAE level 4 systems are actively being deployed and tested on public roadways through fleets of varying size. ADS-equipped vehicle fleets also vary in vehicle platform, operational design domain (ODD), and purpose, with use cases ranging from long-haul trucking to urban ride-hailing. Regardless of the needs being targeted, there is a shared interest by the automotive safety community (e.g., developers, regulators, evaluators, insurers, and academics) to study and quantify the extent to which these deployed fleets have effectively reduced traffic-related fatalities, injuries, and property damage.

There is a longstanding history of "safety impact", sometimes referred to as "safety benefits" or "real-world effectiveness," studies examining the effects on safety of automotive safety interventions like passive and active safety features, vehicle crashworthiness, traffic laws, and, more recently, automation features. The ultimate goal of any safety impact study is to estimate the effectiveness of the introduced countermeasures in reducing crashes that can and do lead to injuries and fatalities. This is accomplished by examining some performance metric relative to a benchmark, which is sometimes also referred to as a baseline. The focus of this paper is on measuring the safety impact of ADS technology, but many of the same principles apply to studying other types of safety interventions such as those pertaining to roadway infrastructure, policies, and education.

These safety impact studies can largely be broken down into two broad categories: (a) prospective and (b) retrospective. Figure 1 shows the relationship between prospective and retrospective safety impact analysis with respect to developmental and deployment milestones for a single system (e.g., an ADS from a single manufacturer or a type of technology from multiple manufacturers). Prospective safety impact studies make predictions about the potential safety benefits of some future fleet (Najm and daSilva, 2000; Yves et al., 2015; ISO T/R 21934-1). Prospective safety impact is helpful early in the development life cycle prior to extensive deployment for an understanding of the ways and extent to which some technology may affect roadway safety. Conversely, retrospective safety impact studies focus on the historical performance of some technology following its deployment. The value of prospective safety impact research becomes more limited as public deployment scales up, and researchers realize the benefits of on-road driving data from which to assist in drawing conclusions. In both prospective and retrospective studies, care must be taken not to generalize the results to other types of technologies or to different manufacturers.

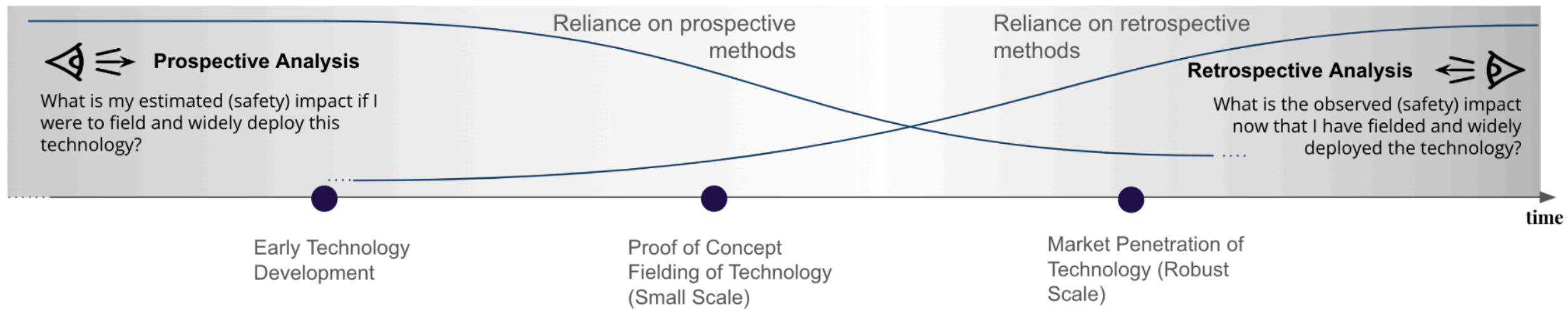

**Figure 1.** Example lifecycle when evaluating a single system using a prospective or retrospectively safety impact analysis.

This paper focuses on retrospective safety impact assessments. This type of analysis has been implemented for measuring the efficacy of many types of safety systems, including airbags, seatbelts, antilock braking systems, electronic stability control, blind spot detection, front crash prevention, lane departure prevention, and backing crash prevention (Glassbrenner and Starnes, 2009; Fildes et al., 2015; Lie et al., 2006; Teoh, 2022; Strandroth et al., 2012; Cicchino, 2017, 2018; Isaksson-Hellman and Lindman, 2015).

Retrospective safety impact is generally presented as a change in safety outcome rate relative to a benchmark rate. Consider this conclusive statement in Cicchino (2017) on real-world performance of Forward Collision Warning (FCW) and Automatic Emergency Braking (AEB): “FCW alone, low-speed AEB, and FCW with AEB reduced rear-end striking crash involvement rates by 27%, 43%, and 50%, respectively.” This statement was made possible by comparing the crash rates (rear-end striking police-reported crash involvements per insured vehicle years) of vehicles equipped with the system being tested (treatment) and the same vehicles without the system as the benchmark (control).

In this paper, attention is focused on establishing best practices for retrospectively evaluating how often crashes occur, generically referred to as “crash rates.” Investigations into crash rates commonly present this rate as a crash event (of some severity) with respect to some measure of exposure, such as vehicle miles traveled (VMT). The primary objective was to prioritize crash events that can (e.g., high risk) or do (e.g., observed outcome) result in occupant injuries, to align with the primary focus of most traffic safety efforts. These recommendations are intended to assist researchers in appropriately and openly comparing crash rates for benchmarking purposes.

**List of RAVE Recommendations**

Table 1 lists out the fifteen recommendations developed by the working group. Specific recommendations are presented in following sections, which are aggregated at the end of the paper in the RAVE (Retrospective Automated Vehicle Evaluation) checklist. The recommendations are broadly grouped into three categories: (1) quality and validity, (2) transparency, and (3) interpretation. Quality and validity covers key methodological considerations for improving study results. Transparency explains some minimum reporting actions that can be taken by researchers to ensure that stakeholders can assess the soundness of the analysis, and thereby establish the reliability of the work. Interpretation covers the fundamental principles of presenting research studies in a way that accurately represents the study design and methodology, while setting boundaries on what can and cannot be inferred from the results.

**Table 1.** RAVE recommendations being put forward by the working group.

| | |
|---|---|
| **Quality and validity** | |
| 1 | Ensure compatibility of the ADS and benchmark data for both crashes and exposure.<br>*There are multiple compatibility challenges when working with ADS and benchmark data sources that require alignment methodology to be employed. This recommendation describes the general notion of performing alignment, some common challenges, and what methodological techniques might be considered for data alignment.* |
| 2 | Prioritize methodological accuracy but default to conservative analysis.<br>*Researchers should strive to align the data and address potential confounders as best as possible to produce unbiased estimates. When addressing remaining potential biases, we recommend that sensitivity analyses should be performed and/or that ADS safety should be underestimated as a conservative approach.* |
| 3 | Favor outcomes that are directly measurable.<br>*The most easily interpreted outcomes in any observational study are the ones that are directly measurable.* |
| 4 | Consider outcome data transformations to help overcome limitations and address new research questions.<br>*When directly measurable data provides limited meaning, transformations can help alleviate data limitations or even help answer different sets of research questions.* |
| 5 | Quantify uncertainty of the estimates with statistical testing or other methods.<br>*Statistical conclusions are a fundamental result presented in safety impact studies and researchers should catalog and quantify sources of uncertainty.* |
| **Transparency** | |
| 6 | Cite all data sources used in the study.<br>*A wide range of data sources will be relied upon for evaluating ADS safety. Transparency in reporting what sources are used as well as unique strengths, weaknesses, or peculiarities of the data is important for assessing the accuracy and credibility of the study.* |
| 7 | Provide descriptive statistics of the ADS and benchmark data composition.<br>*A descriptive summary provides background information on key variables, as well as unweighted and weighted distributions, broken down by each group (ADS and benchmark), to indicate sources of potential bias that can potentially compromise results.* |
| 8 | Describe the ADS deployment being evaluated.<br>*ADS technology is rapidly evolving. Researchers should precisely describe the operating and deployment characteristics of the automated vehicles in the study sample and cite supporting information when possible.* |
| 9 | Clearly document analysis decisions and steps in the methods section.<br>*Methodological steps need to be carefully described in a way that would allow someone with access to the exact same data to replicate the study results and for stakeholders reading study documentation to understand and be able to reflect on what the implications are of the specific methodological decisions and steps.* |
| 10 | Document limitations in the study's data, methods, scope, and interpretation.<br>*Every study has limitations, typically resulting from the selected data sources, data collected, and analytical decisions made. Documenting study limitations should be embraced as an opportunity for researchers to transparently communicate the strengths and weaknesses of their study.* |
| **Interpretation** | |
| 11 | Build upon past research and justify the scope for what was selected to be studied.<br>*Existing literature should be considered and discussed when presenting study motivation, design choices, and results. Researchers should discuss and justify the choices of which analyses were and were not performed.* |
| 12 | Develop research questions that are clear, concise, and specific.<br>*Research questions should be clearly stated with a level of clarity that translates to a foundation for the methodological decisions, results, and conclusions to follow.* |
| 13 | Ensure conclusions accurately reflect the study design.<br>*It is important to restrict conclusions to those which follow logically from the research question, accurately represent the data and methodology being relied upon, and consider the study's limitations.* |
| 14 | Exercise caution in relating findings between different severity levels.<br>*Lower severity crashes happen more frequently than higher severity crashes, but the contributing factors in low severity crashes are not necessarily common in high severity crashes. Extrapolating lower severity crash performance toward high severity performance should be done cautiously, and accompanied by discussion of how contributing factors were accounted for and of the limitations in such an extrapolation.* |
| 15 | Present rates in incidents per exposure units.<br>*Rates presented as incidents per exposure have the distinct advantage of being linearly related to the expected count of incident outcomes, which improves both interpretability and comparability.* |

## QUALITY AND VALIDITY

Study quality and validity can vary between research efforts as a direct result of the data relied upon and methodological choices being made. Beginning with the formulated research question, researchers need to make choices about data selection, data processing, and statistical approach. These methodological choices directly alter the results, which, if not properly contextualized, could lead to erroneous conclusions. This section describes recommendations on methodologies that increase quality and validity of studies. These recommendations cover data alignment, exposure matching, research question development, conservative analysis, outcome selection, and quantification of uncertainty.

### Ensure data alignment in ADS and benchmark data for both crashes and exposure.

A large part of retrospective safety impact methodology centers around data alignment procedures. Because many ADS and human data sources are reported separately, some amount of data alignment is required in order to make legitimate comparisons. Crash and exposure (e.g., VMT) data can also be available from separate sources, further complicating alignment. Common comparability issues that should be addressed in data alignment are discussed in the subsections below and include:

1. Differing reporting practices between ADS and benchmark data sources for both crashes and exposure,
2. Varying exposure to influential factors, and
3. Mismatched outcome and exposure units.

There are a wide range of analysis methods that might be considered to align the data sources, some of which are listed in Table 5. The required methods are dependent on the data sources being used and the intended research questions, and a given study may require multiple methodologies.

**Table 5.** Methods for aligning ADS and benchmark data.

| Method | Description | Usage |
|---|---|---|
| *Subselection* | Subsetting data using inclusion criteria | Commonly relied upon in observational research to isolate control and treatment data that were collected under more comparable conditions (Portney and Watkins, 2009) |
| *Imputation* | Filling in missing data | Incomplete data is commonplace in crash data, where only partial evidence is recorded following an event. These missing data elements can be approximated using a variety of approaches (e.g., Herbert, 2019). |
| *Transformation* | Converting raw data into a processed form | Raw data generally has limited utility in its original form. Simple transformations, like reducing data using a classification routine, can help facilitate analyses. Re-weighting schemes, such as underreporting adjustments or injury risk scoring, are other examples. Modeling techniques can also be useful, whereby the raw data is used to build some model of crash rates (discussed in "Favor outcomes that are directly measurable"). |
| *Joining* | Combining features from multiple data sets | Individual data sources are often limited in what they record. Combining multiple data sets with complementary values is often a requirement for analysis. |

**Differing reporting practices:** Crash reporting differences are one of the biggest challenges in comparing ADS to human benchmarks, and must be addressed in data alignment procedures. There are a wide range of crash reporting thresholds being used in both ADS and benchmarking databases, and datasets are expected to evolve over time. As an example of this challenge, NHTSA issued Standing General Order (SGO) 2021-01, which set reporting requirements for all ADS deployments in the United States (NHTSA, 2023). NHTSA publishes incident reports generated from this SGO in a centralized location, and this publicly-accessible dataset has served as the source of ADS crash data in multiple safety impact studies (Chen and Shladover, 2024; Cummings, 2023; Kusano et al., 2024). The SGO currently requires all crashes be reported that result in "any property damage, injury, or fatality," where the ADS was either involved in or contributed to (or alleged to contribute to) a crash. This ADS SGO reporting requirement does not align with available human crash databases, which include a wide range of police reported data, insurance data, and naturalistic driving studies (NDS). For example, national estimates indicate that fewer than half of humans' crashes, where either an injury occurred or there was some amount of property damage, are reported to police (Blincoe et al., 2023). While some of these may not meet minimum property damage thresholds for police-reporting, about a third of injury crashes are also unreported.

Reporting differences, however, do not directly preclude using ADS and benchmark data in a retrospective safety impact analysis. Rather, researchers should acknowledge any potential data limitations, clearly communicate methodological actions to mitigate such limitations, and properly contextualize results. One straightforward alignment technique may be to subselect each dataset to minimize reporting differences and any data underreporting concerns. For example, objective measures, such as filtering using the occurrence of airbag deployments or setting some delta-v threshold (for vehicle-to-vehicle and single vehicle crashes), might be helpful in some instances to more closely align crash inclusion criteria. As an example, event data recorders (EDR) that are Part 563 compliant have a required non-deployment triggering threshold of 5-mph delta-v, and there has been early work to use this triggering threshold in generating benchmarks (NHTSA 49 CFR Part 563; Flannagan et al., 2023). When such objective measures are unavailable, another approach used by Teoh and Kidd (2017) is to code crash incident narratives and make a subjective determination of whether such an event would typically have been reported to police under different circumstances (i.e., not an ADS vehicle operated by a well-known company). This largely involves determining if the crash damage clearly exceeded the police reporting minimum level (police-reportable). Some subjective evaluation was still performed to determine whether the nature of the crash would have elicited a call to police in the first place (addressing underreporting to police). Rather than downsampling ADS crashes, another methodology that has been explored is the reliance on adjustment factors to upweight human benchmark data to account for differences in reporting thresholds and crash underreporting practices (Bärgman et al., 2024; Blincoe et al., 2023; Blanco et al., 2016; Schoettle and Sivak, 2016; Scanlon et al., 2024).

**Varying exposure to influential factors:** Many environmental, driver, vehicle, and occupant factors can influence crash risk. This might include factors like geographic region, road types, speed limit, vehicle type, time-of-day, and weather (Choudhary et al., 2018; Farmer, 2017; Kmet and Macarthur, 2006; Kweon and Kockelman, 2005; Martin, 2002; Regev et al., 2018; Qiu and Nixon, 2008; Rice et al., 2003). Given the research question that has been set forth, researchers must consider how confounding factors may influence results, conclusions, and, ultimately, interpretation of the study. Three components should be considered in any analysis:

1. Identify potentially influential factors.
2. Consider differences in exposure and effect size.
3. Execute a mitigation strategy.

What factors may confound results? Specifically, what conditions of the ADS operating environment may decrease or increase crash risk? The most straightforward way to identify these factors may be to simply consult the longstanding traffic safety literature and consider how each feature may influence the operations being evaluated.

Both the differences in exposure and the effect size will dictate how important it may be to account for a confounding factor. As discussed in a later section, a "Table 1" is often used in epidemiology studies and provides a useful means for transparently showing differences in exposure between the ADS and benchmark population. The importance of accounting for a specific factor can differ between analyses. For example, a factor known to affect crash risk may be less important to account for in a particular study if the exposure is similar between the two comparison groups.

There are multiple mitigation strategies that a researcher may take when differences in exposure to influential factors exist. Each option centers around the idea of promoting "apples-to-apples" comparisons. Consider a scenario where an ADS drives only on surface streets, but benchmark data may be available for all road types, including high speed freeways. One option available is to *subselect* the benchmark driving data to achieve similar populations. In this example, one may choose to only analyze benchmark data from surface streets. Now consider a scenario where the ADS driving data has some small amount of freeway exposure that is far less than the proportions in the benchmark population. A second option available may be to match the driving exposures to each other. For example, using the observed benchmark crash rates on each road type, researchers could determine the expected crash count for the benchmark population if they had a similar driving exposure to the ADS. Another option for this hypothetical would be to assess the driving, independently, in both driving environments as is discussed in the next section.

**Mismatched outcome and exposure units:** Outcome and exposure units must also be matched. Comparing a rate of crashes per million miles to crashes per million kilometers would, of course, need a unit conversion to make the two values comparable. There is, however, a more nuanced unit error that has been made in computing crash rates, and that is the distinction between a vehicle-level and a crash-level crash rate (Scanlon et al., 2024). ADS crash rates have exclusively, to-date, been presented as a vehicle-level crash rate, where the vehicle travels some amount of distance and experiences some number of crash events. Where incorrect, the number of times a crash event occurred was counted for a population of drivers, where a single crash might involve multiple involved vehicles. To correct this approach and convert to a vehicle-level rate, the total number of involved vehicles (or drivers) must be counted instead (Teoh and Kidd, 2017; Lindman et al., 2017; Scanlon et al., 2024). Depending on the severity level, most crashes involve more than one vehicle, where, nationally, a vehicle-level rate has been shown to be 1.8 times that of a crash-level rate (Scanlon et al., 2024).

**Prioritize methodological accuracy but default to conservative analysis.**

Following the previous recommendations, researchers should strive to align the data and address potential confounders as well as possible to produce unbiased estimates. Even the best efforts and available data may not be able to account for all important covariates. When there are likely to be some biases remaining, we recommend that researchers should perform sensitivity analyses and/or underestimate ADS safety as a conservative approach. The essence of this recommendation is that (1) no study is perfect given the intrinsic incompleteness of the problem and (2) at times the highest accuracy approach is unclear or infeasible. Safety impact studies are performed with a research question in mind and are motivated by a need to inform a specific or general set of stakeholders. If a non-conservative approach is chosen, such a stakeholder may erroneously get the impression that a specific ADS is safer than it actually is. By selecting a conservative approach, the same stakeholder would get "at least" the performance provided to them by the evaluation.

Sensitivity analyses, in particular, can show the differences among diverse analytical strategies. Consider the study by Kusano et al. (2024) comparing Waymo ADS rider-only data to police-reported benchmarks. They aimed to evaluate rates of "any property damage or injury" using ADS SGO data and an underreported adjusted benchmark. Even with the adjustment, there was still uncertainty as to what severity level, from the ADS side, may be considered equivalent to this adjusted benchmark. To explore the effect of this methodological decision, a secondary analysis subsetting out ADS data below a 1-mph threshold was employed to evaluate how dependent the results were to cases with very low collision impulses. This sensitivity analysis demonstrated that about half (48%) of the ADS crashes had delta-v estimates less than 1-mph, which demonstrates just how sensitive the results are to lower reporting threshold. Another type of sensitivity analysis is to bracket estimates using a range of possible values. For example, Blanco et al. (2016) used a "low", "medium", and "high" property damage only crash underreporting correction to police-reported data to explore the sensitivity of the underreporting adjustment on the comparison to an ADS. Similarly, Kusano et al. (2024) presented a comparison of an ADS to an observed (uncorrected) any-injury-reported crash rate and an underreporting adjusted benchmark. It is known there is some amount of underreporting in human data (Dingus et al., 2006) that is also likely higher than the underreporting in the ADS data. However, the true underreported crash rates were unknown and the adjustment methods were approximations, so including results with (an estimate) and without (a lower bound) this adjustment enables stakeholders to understand the statistical effect of this data transformation methodology.

Alternatively, conservative assumptions that underestimate ADS effectiveness can be used. An example of a conservative assumption from Kusano et al. (2024) was to assume any ADS crash with the "Law Enforcement Investigating?" field reported as "Yes" or "Unknown" in the NHTSA SGO data was police-reported and thus comparable to state and national police-reported crash databases. The "Law Enforcement Investigating?" field indicates whether law enforcement was investigating (e.g., arrived at the scene and took information, and/or indicated that a report would be filed). It is not possible to verify that a police report is ultimately filed and further if the police report is included in the state police accident report databases, which are often published months to years after the calendar year ends. The conservative assumption to include any ADS crash with the corresponding SGO field was chosen by researchers because there was no feasible way to investigate the sensitivity of this police-reporting practice in the timeframe of the study.

**Favor outcomes that are directly measurable.**

The most easily interpreted outcomes in any observational study are the ones that are directly measurable from on-road driving events. Directly measurable refers to those features that can be observed or measured (raw data) and do not require processing, such as through the data transformations discussed in the following section. Directly measurable outcomes might include, for example, crashes where an injury outcome occurred above some predefined threshold, such as using the often encoded KABCO scale to identify crashes with a fatality (designated using “K”). Similarly, directly measurable quantities or events that are correlated or associated with injury outcomes, such as EDR-measured delta-v or airbag deployment occurrence, can be useful.

Sometimes, directly measurable data suffer from biases. Reporting practices are one area where bias can be introduced. Consider a comparison between ADS and human-driven vehicle tow-away crashes, where a human-driven vehicle is often towed following a crash due to damage that compromises the vehicle's ability to be driven. Conversely, ADS-equipped vehicles are often towed following crashes with relatively lower forces due to external sensing capabilities having been compromised. Compromised sensing capabilities have the potential to disrupt the ADS’s operability, which could require the vehicle to be towed. Whereas, under similar collision conditions, a human-driven vehicle may not have needed to be towed. Further, company-specific operational policies that dictate a location of deployment may lead to disproportionate ADS towing with respect to a benchmark population.

**Consider outcome data transformations to help overcome limitations and address new research questions.**

When raw, directly measurable data provide limited meaning, transformations, such as underreporting corrections or risk-based transformations, can help alleviate data limitations or even help answer different sets of research questions. While, directly measurable data has the advantage of not being influenced by transformation techniques that alter the raw data, there are a wide range of reasons why researchers might look to transform the outcome.

There are a wide range of reasons why researchers may look to transform outcome data. The aforementioned underreporting adjustments are one example, where raw crash counts for lower severity crashes are reweighted to account for missing events that would have otherwise qualified (Blincoe et al., 2023; Blanco et al., 2016; Schoettle and Sivak, 2016; Scanlon et al., 2024). This transformation, although useful for aligning ADS and benchmark data, introduces new potential inaccuracies and imprecision. In these examples, data subselection permitting, choosing to also focus on higher severity crashes that do not suffer the same underreporting concerns would serve to supplement the “adjusted” crash rates analysis and improve analysis credibility. As deployment expansions continue and the opportunity arises to evaluate ADS safety impact in rarer, higher severity outcomes, it is expected that analysts may shift their focus away from minor to moderate injury causing collisions and toward serious and fatal crashes (AAAM, 2015).

While observed injury outcomes can be influenced by occupant considerations (e.g., number, age, seating position) and some associated randomness, metrics transforming the outcome data using risk-based approaches may have merit. Variations of this well-established "dose-response" safety impact approach assign an injury risk-based value to each event using the dynamics of the crash and established injury risk curves (Kononen et al., 2011; Kullgren, 2008; McMurry et al., 2021; Schubert et al., 2023, 2024; Lubbe et al., 2022). This metric can then be used to investigate the propensity of some collision conditions to elicit some injury outcome (Kusano and Gabler, 2012; Kusano et al., 2024; Scanlon et al., 2017, 2022). A simple form of this risk-based approach may categorize severity level according to injury risk, such as Flannagan et al.'s (2023) "meaningful injury risk" based delta-v measures or S-level severity bucketing in ISO 26262 based on probability of a Maximum Abbreviated Injury Scale (MAIS) injury in an event (ISO 26262-1:2018). If injury risk functions are used, the authors should provide enough details so the injury probabilities can be recreated (e.g., citations to which risk functions were used, injury risk function model parameters used or inferred, and intermediate values such as delta-V or collision speeds used to compute injury risk). Caution should be used and limitations should be stated when relying on risk-based approaches, however, as there may be unaccounted for assumptions around seating positions, belt status, vehicle structure design, passive safety features, multiple events, specific crash types, among others. Some studies have demonstrated that reporting thresholds are needed when transforming crash data using injury risk curves due to the potential for unreported events and the tendency for injury risk curves to overestimate true injury risk at low collision severities (Bärgman et al., 2024). A potential use of injury risk functions may be to set objective inclusion criteria that can be equitably applied to both benchmark and ADS data. For example, the analysis decision to select only vehicle-to-vehicle crashes that have a maximum crash delta-V above some threshold could use an injury risk curve to justify the injury-relevance of the delta-V threshold.

While potentially useful, as described above, transformation techniques are a form of extrapolation that can potentially introduce uncertainty into the analysis and potentially increase (or decrease) bias. It is recommended that, if any transformations are applied, researchers document and account for any bias and uncertainty that was potentially introduced.

**Quantify uncertainty of the estimates with statistical testing or other methods.**

It is best practice to leverage statistical methods to quantify uncertainty in retrospective safety impact analyses. There are a wide variety of methods that can be used in this context, ranging from normal/Poisson confidence intervals to various regression models that account for covariates and other issues. Researchers should use appropriate statistical methods to report standard errors and/or confidence intervals for any descriptive statistics or crash rate comparisons presented. The use of statistical testing accompanied with thorough documentation about the appropriateness of using the particular test or distribution choice given the data (e.g., related to normality or other methodological constraints) is suggested. For uncertainty in elements for which distributions are not known, sensitivity analysis and bootstrapping approaches can be useful. It is also important to describe other sources of uncertainty that were not accounted for in the study and their potential effect on the results.

## TRANSPARENCY

A goal of any scientific writing is to provide transparency in the data, methodology, and inferences that enable readers to, in principle, replicate the study or draw their own conclusions. With that being said, researchers are faced with numerous decision points when designing and executing a research study (Gellman and Loken, 2013). There are also challenges with using ADS and benchmark data that is not always publicly available, which can make it harder to achieve scientific reporting that truly supports replication. Transparency not only fosters trust in this new, challenging science, but helps facilitate future research that incrementally increases the body of knowledge. In this section, we highlight specific areas where communication is especially critical in this context.

**Cite all data sources used in the study.**

There are multiple data sources that researchers can use to evaluate the safety of automated vehicles. Sander et al. (2024) provides a thorough review of a variety of existing benchmark data sources. Rather than enumerate the numerous government-managed, naturalistic, and insurance data sources, it is important to acknowledge that a wide range of data sources may be relied upon for evaluating ADS safety impact, where each source will have its own unique value and limitations.

Nevertheless, data used to compute crash rates or other metrics must be described in detail so the research can be critically evaluated by peers, replicated, and extended. At a minimum, studies should include the name of the data source, the origin of the source, the date range of information used, the sampling or reporting frequency (e.g., annual vs. monthly), and references to any relevant documentation detailing the data source (e.g., data dictionaries). It is notable that not all data relied upon in analyses will be publicly available, which makes it harder to achieve scientific reporting that truly supports replication. Nevertheless, transparency in reporting about any data relied upon is important for furthering credibility.

Researchers also should report other peculiarities about the information or records contained in the data source. This might include indicating any sampling scheme (if used), inclusion and exclusion criteria, and data reporting considerations. This information can be communicated, for example, in a single table with supporting text (e.g., Teoh and Kidd, 2017), a diagram or other illustration, or as detailed descriptive text.

**Provide a descriptive summary of the ADS and benchmark data composition.**

We suggest that researchers provide descriptive statistics of the ADS and benchmark data, which are commonly presented in a table form. These sort of descriptive tables are colloquially referred to as a "Table 1" in the field of epidemiology, where they are commonplace (Hayes-Larson et al., 2019), and have also been used in evaluating crash risk (Choo et al., 2023). The role of Table 1 is to provide background information on key variables broken down by each group (e.g., ADS and human-driven vehicles), to indicate sources of potential bias that can potentially compromise results (Hayes-Larson et al., 2019). Any such table should include descriptions about both the raw data form and any sort of subselections and transformations applied. Researchers using the same data may make different decisions or use different approaches to estimate the safety impact of ADS, possibly based on different research questions, so the data ultimately used in the analysis must be presented in a way that is traceable back to the raw data.

How each characteristic is described in the descriptive summary table will vary by data type. Counts and proportions are commonly used for categorical variables. Measures of central tendency and variance are commonly used to describe continuous variables. For further guidance on design of summary tables, researchers can refer to Hayes-Larson et al. (2019).

Some generally used characteristics, however, are the number of vehicles or observations, the outcome variable and its components (e.g., crash rate, number of crashes), and the exposure measure (e.g., number of miles traveled, insured vehicle years). Researchers also should consider including characteristics that describe the sample data and cohorts as a function of study design that is disaggregated by relevant and influential variables. As an example, human driver characteristics influencing crash risk like age, gender, and driving experience can be used to characterize a human driver benchmark group (Ivers al., 2009; Rhodes and Pivik, 2011). Other information like geographic area, road type, lighting (e.g., day, night), or posted speed limit can provide more detailed information about exposure for the entire sample and separately for the experimental group and the control group (Rice et al., 2003, Farmer, 2017). Potential biases in the data are typically clear and evident once such disaggregation occurs, and can subsequently be addressed or treated as limitations.

**Describe the ADS deployment being evaluated.**

ADS technology is rapidly evolving. Researchers must precisely describe the operating and deployment characteristics of the ADS in the study sample and cite supporting information when possible. Failure to contextualize the unique ADS deployment conditions (at a particular point in time) could lead to misinterpretation or misapplication of findings. The description of the ADS deployment should be precise enough to allow future research to aggregate or synthesize findings across multiple related safety impact studies.

The ADS deployment description may include, but would not be limited to, the automated vehicle company, study sample time range, ODD, vehicle platform, driving location(s), presence of a human operator or supervisor, and the type of operation (e.g., ride hailing, long haul shipping, or local delivery). Some features to consider when describing the ODD and characterizing the ADS deployment can be found in multiple standards (AVSC, 2022; BSI 2020). Important features of the ADS deployment should be included in the descriptive summary table to highlight similarities or differences between the ADS data and the benchmark data.

**Clearly document analysis decisions and steps in the methods section.**

There are many methodological approaches that researchers may choose to follow in their analyses. We recommend that any methodological steps need to be carefully described in a way that would allow someone with access to the exact same data to replicate the study results. This careful description is also important for stakeholders reading study documentation to understand and be able to reflect on what the implications are of the specific methodological decisions and steps on the assessment results and their implications. This includes providing references to any former methodology which the current study is using directly or expanding upon. If additional annotations and classifications are applied to the data, there should be corresponding text defining how they were added.

As an example, researchers might consider documenting both the variable names and their respective values relied upon in any classification routines, where these variable and value pairings can be used to produce exact replication of results (e.g., which vehicle body type codings were considered passenger vehicles?). If manual case review was performed to generate previously unavailable variables, a data table for joining the supplemental data with the main dataset(s) may be helpful. Alternatively, researchers may even consider providing source or pseudo code for any data transformations. Providing code is a requirement of some journals and is considered best practice. Doing so clarifies exactly what was done and supports replication or, at least, transparency when the data themselves cannot be shared.

**Document important limitations in the study's data, methods, scope, and interpretation.**

Research manuscripts should include a limitations section where authors describe confounds, sources and effects of bias, issues with generalizability, ethical considerations, and other factors that may have influenced the study results and conclusions. Every study has limitations, often resulting from the selected data sources, data collected, and methodological decisions made. Describing study limitations should be embraced as an opportunity for researchers to transparently communicate the strengths and weaknesses of their study. If possible, researchers should predict the direction and/or magnitude of any possible bias created by these limitations or methodological decisions. Limitations often lead to additional research questions to address in future studies, which will effectively serve to strengthen the understanding of a topic and complement the state-of-the-art. Importantly, being transparent about study limitations increases credibility and reputation.

Safety impact studies that compare ADS safety with a benchmark should be comprehensive in describing the limitations of their study. A common limitation of ADS safety impact studies is the assessment scope, since most studies only examine the deployment experience of one or a few ADS companies over a limited study duration and ODD. Documenting this limitation helps inform stakeholders of the ADS assessment scope being evaluated and to caution against extrapolating the results toward some out-of-scope assessment. Similarly, the findings will be limited to a subset of crashes due to data restrictions or methodological choices that confine the data to certain crash types, severity levels, roadways, or other characteristics related to data reporting and exposure. Laws and regulations (e.g., crash reporting requirements and permitting), particularly if they change during the study period, can also limit the validity of the study findings. See the recommendations under "Quality and Validity" on methods that should be used to limit the effects of known limitations when comparing ADS data to benchmarks.

**INTERPRETATION**

Researchers have a responsibility to clearly present their methodological findings in a way that accurately represents the study design and results. Similarly, study evaluators (e.g., paper reviewers) have a responsibility to meticulously examine the presented conclusions in light of the existing state-of-the-art and inherent study limitations. Not all readers of a study will be subject matter experts. Although on its face relatively simplistic, small differences in retrospective safety impact methodologies can dramatically alter findings and nuances in descriptions can alter interpretation. Within this paper, a set of basic recommendations are proposed that are considered to be important elements for facilitating accurate study interpretation. We recognize that the content of this section largely reflects basic scientific practice, but feel that an adherence to this guidance can greatly enhance interpretability of the implications of individual works.

**Build upon past research and justify the scope for what was selected to be studied.**

The design of any study on ADS retrospective safety assessment should begin with thoughtful consideration of past works. If the researcher aspires to fill gaps in the body of knowledge, this preparation enables them to identify the gaps. In any situation, consideration of past work provides the researcher with lessons learned to inform a study's design, objectives, and interpretations, even if a planned hypothesis is opposite what is found in previous work. Consideration of previous work should be clearly documented in the publishing of any new study.

Such incorporation of previous work provides stakeholders with proximate reference points against which they can objectively evaluate a study's methods and findings.

Researchers should discuss and justify the choices of which analyses were and were not performed. Consider examining effects of your treatment on various subsets of the data, such as different crash types, severity levels, geographical locations, and road types. Multiple independent analyses looking at the data from different perspectives can serve to strengthen conclusions. Considering the wide range of research questions may help to inform stakeholders and further broaden the methodological approaches. Practically, however, additional analytical lenses means splitting the data into smaller analysis units, which may reduce statistical power in detecting differences between ADS and benchmark data. If smaller subsets of the crash population are evaluated, researchers should discuss and justify the choices of which subsets to report and not report. For example, if a study reports on crash rates in intersection crashes during the day, but excludes other crash modes and times of day, the study authors should discuss why the other modes and times of day were excluded from the data presented (e.g., the ADS only operates during the day so other times of day are not relevant; see the section on data alignment).

No study is perfect, and not all studies' results can be combined directly. However, alignment on high-level conclusions across multiple studies (e.g., observing benefits; using a variety of methods and/or study populations) can be used as an indication of stronger evidence. Similarly, one study's findings should not be interpreted in isolation, but rather in consideration of and within the context of the existing body of knowledge.

**Develop research questions that are clear, concise, and specific.**

Research questions, sometimes referred to as evaluation scope (Fahrenkrog et al., 2024), are fundamental to an entire study process, from design to interpretation, and thus should be stated early, clearly, and concisely. An insufficiently scoped research question can lead to imprecise methods during study execution and difficulty in interpretation of study results. As an example, a research question "Is a particular ADS safer than human drivers?," while clear and concise, is overly broad and does not provide a clear indication of what systems are being evaluated, what outcomes are being measured, what benchmark population is being used, and any scoping restrictions. On the other hand, the research question "Does a particular rider-only ADS system have lower rates of police-reportable crashes than human drivers within the same geographic area?" has specificity that tells the reader exactly what to expect to be answered by the study. It also tells the reader what not to expect. For instance, absent any other research questions, the reader knows this is not a study about safety in terms of number of disengagements, near-crashes, simulated outcomes, and so on.

Research questions should be stated only with a level of specificity that is warranted given the data being relied upon and any methodological procedures being used. Researchers should also explain how the existing body of knowledge and any novel information motivated their research question(s). Clearly defined research questions also help readers understand how to compare the results of different studies, which may have slightly different contexts, data, and objectives. Alternatively, the research question can indicate that the results are not comparable at all. A lack of clarity in the research question(s) translates to uncertainty for the study reviewer when evaluating the research. This lack of clarity may also be a reflection of the research itself, in that it was not properly scoped, and is attempting to draw conclusions not supported by the methods and results.

**Ensure conclusions accurately reflect the study design.**

Conclusions should follow logically from the research question, accurately represent the data and methodology being relied upon, and consider the study's limitations. As discussed above, the research question establishes the scope of the study. Analyses have historically studied the performance of one or several ADSs at some defined outcome levels in some operational design domain (ODD), where ADS performance is generally limited to specific vehicle body types, roadway types, and geographic areas. The results should not, in turn, be used to make more generalized conclusions outside of this scope. For example, it would be inappropriate to make conclusions about a reduction in fatalities from a study that is restricted to only looking at any-injury-level crash rates, where the former was not within the evaluation scope and was not investigated as a part of the study design. In another example, early studies (Blanco et al., 2016, Goodall et al., 2021a, Banerjee et al., 2018, Dixit et al., 2016, Favarò et al., 2017, Schoettle and Sivak, 2015, Blanco et al., 2016, Teoh and Kidd, 2017.) showed promising safety results for ADS in testing operations with human supervision behind the wheel. At least two studies have included a mix of ADS driving without a human behind the wheel with rider-only operations (Cummings, 2023; Di Lillo et al., 2023). These prior studies investigating performance with a person behind the wheel provide limited safety impact signals for rider-only ADS performance but can serve to help motivate future work. Because the human can disengage the vehicle, it would be inappropriate to make retrospective claims about rider-only operations, where there is no human supervision and the technology and deployment area are radically different.

In addition to considering the results, a study's conclusion must consider its limitations and its sensitivity to limitations. Would resolving a limitation be expected to undermine the conclusions? If a set of limitations is so influential that it likely compromises the integrity of any results, drawing conclusions should be done with care or, possibly, results should be deemed inconclusive. This is especially true when a specific limitation has been found to be highly influential in the existing literature. Studies must clearly state their limitations alongside any conclusions to communicate the strength of the conclusions to readers, which include non-experts.

**Exercise caution in relating findings between different severity levels**

In performing safety impact related analyses, more driving exposure (e.g., distance traveled) is required to draw statistically significant conclusions for higher severity crash outcomes than for lower severity outcomes (Kalra & Paddock, 2016; Scanlon et al., 2024). Lower severity crashes, such as property damage only crashes, happen more frequently than higher severity crashes, such as those resulting in serious injury or fatality. For researchers planning a safety impact analysis, when faced with the low counts of higher severity crashes, it is often tempting to emphasize analysis of lower severity crashes due to the increased statistical power they provide through greater availability of observations. This approach is not without merit, but care should be taken when attempting to extrapolate those low severity results toward higher severity outcomes. Low severity crashes (and all severity level crash collections) can share injury mechanisms or contributing factors with higher severity crashes (Kusano et al., 2023; Lindman et al., 2017). For this reason, analysis of available low severity events can provide some guidance. However, events of different severity categories will have their own specific set of contributing factors, which is sometimes referred to as vertical heterogeneity (Knipling, 2017; Kusano et al., 2023). For example, implementing countermeasures for factors in low severity crashes could help reduce high severity crashes, but may not address all factors in high severity crashes, or address factors or injury mechanisms in the same proportion as would be addressed for the observed low severity crashes (an X% reduction in low severity outcomes does not imply an X% reduction in high severity outcomes). This consideration is particularly important when evaluating the contribution of ADS-equipped vehicles to a Safe System or Vision Zero goal of eliminating serious and fatal injury. These approaches highlight the systems perspective and interaction of multiple factors beyond the driver, driving system, or ego vehicle. Examples of these additional factors include the associated roadway design, national legislation, or interpretation of courts (Lie & Tingvall, 2024). While sufficient mileage is accumulated for a system, researchers faced with a scarcity of severe events should acknowledge the challenges to making conclusions beyond the available data and look to leverage prospective methods, including use of appropriate safety surrogates or methods which can directly consider the underlying causal mechanisms in higher severity outcome events.

**Present rates in incidents per exposure units**

A collision rate can either be presented as incidents (e.g., crashes) per exposure unit (e.g., miles) or, its reciprocal, exposure units per incident. We recommend that event rates be presented in incidents per exposure units for retrospective safety analyses. Rates presented as incidents per exposure have the distinct advantage of being linearly related to the expected count of incident outcomes, which improves both interpretability and comparability. The reciprocal rates, exposure units per incident, has a curvilinear relationship. This difficulty has also been noted with fuel efficiency ratings for vehicles, for example the contrast when reporting fuel economy in miles per gallon (exposure unit per incident) versus gallons per mile (incidents per exposure) (Larrick and Soll, 2008).

The non-linear reciprocal rate, exposure unit per incident, creates an opportunity for misinterpretation of relative differences between rates, because there is a tendency for readers to expect that the rates are linearly related when they are not. Specifically, the rate differences are expected to be linearly related to the expected number of outcome events. Several of the authors have anecdotally experienced this in their own work, but, notably, this illusion has been demonstrated in participants interpreting fuel efficiency measurements (Larrick and Soll, 2008; Rowan et al, 2010). This illusion is depicted in Figure 2 for a foreseeable range of benchmark crash rates (e.g., Chen and Shladover, 2024 and Scanlon et al., 2024). Consider one ADS that has a miles per incident rate of 1 million miles per crash compared to a benchmark of 750,000 miles per crash. Another ADS has a 500,000 miles per crash rate compared to a benchmark of 250,000 miles per crash. In both instances, the difference in miles driven per crash is 250,000, giving the illusion that the difference in performance is similar. Contrary to this, the former comparison shows an ADS that reduces the number of crashes per mile by 25% (1 IPMM vs 1.33 IPMM), while the latter reduces the number of crashes per mile by 50% (2 IPMM vs 4 IPMM). Because the incidents per exposure units rates are linearly proportional to the number of events and the exposure unit per incident rates are not linearly related, it is not readily apparent that the relative rates are more difficult to compare.

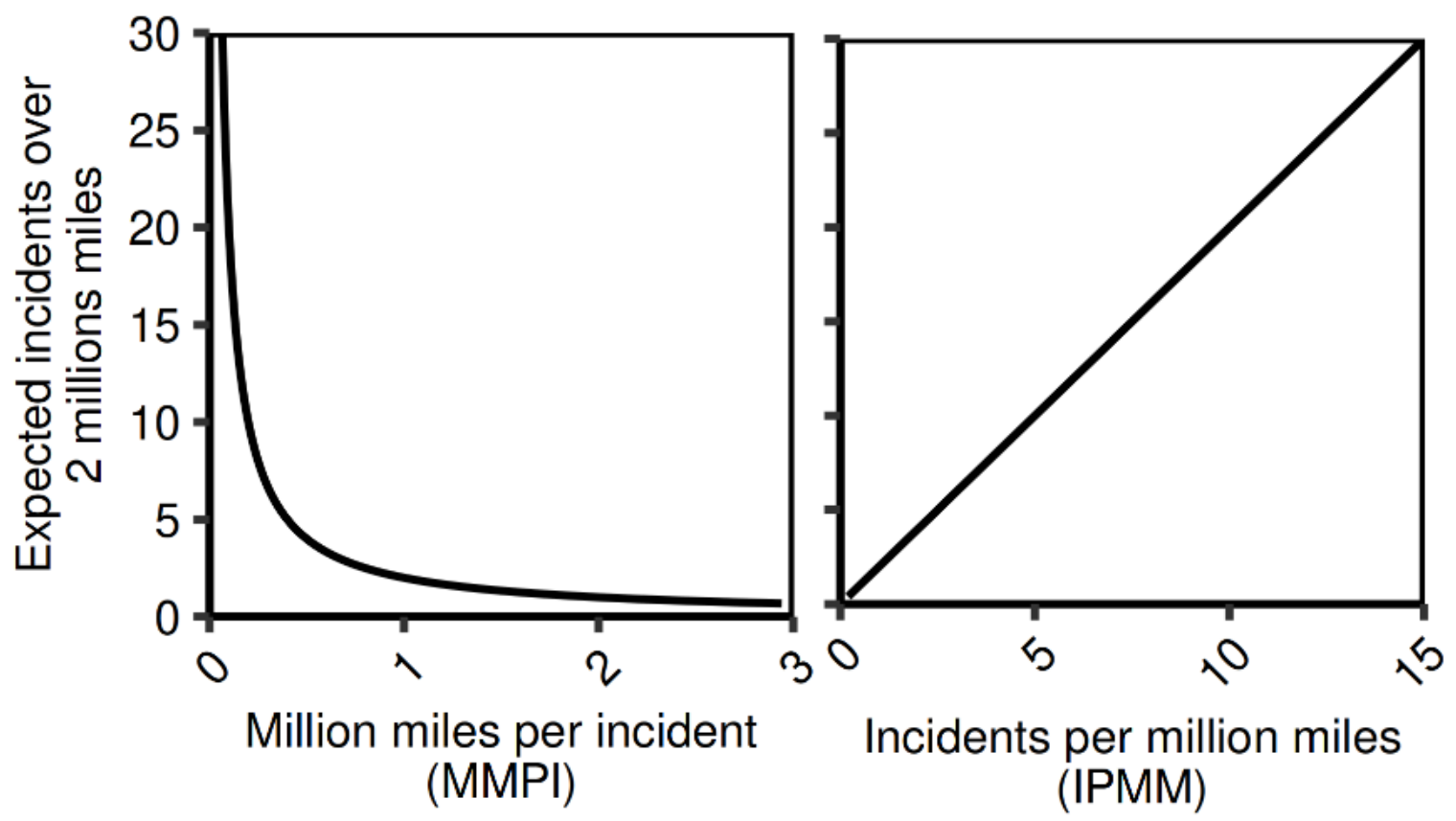

**Figure 2.** The relationship between expected incidents over 2 million miles of driving and rates of million miles per incident (left) and incidents per million miles (right). As the rate measure increases, the curvilinear (left) and linear (right) relationship can be seen for each rate type.

Another key advantage of presenting a collision rate in incidents per exposure is that these rates can be directly added together. For example, a benchmark that has a rate of events of 5 level-1 incidents per mile and a rate of 10 level-2 incidents per mile has a combined level 1 and 2 event rate of 15 (10 level-1 + 5 level-2) incidents per mile. If these same rates were presented as exposure units per incident, the addition is not as straightforward: the rate of level-1 events of 0.2 mile per incident and level-2 incidents of 0.1 miles per incident results in a combined level 1 and 2 rate of 0.067 miles per incident (1 / (5 + 10)).

**RAVE CHECKLIST**

From the aforementioned sections, the working group has developed a list of specific recommendations that can be used as a checklist for researchers performing retrospective safety impact studies as well as those tasked with evaluating them. These specific recommendations are listed in Tables 2-4. Each specific recommendation is either "required" (should be done if applicable and/or if data allows), "recommended" (could be done if applicable and/or if data allows), or "general guidance" (general good practice that may improve the quality of the study).

**Table 2.** Quality and validity specific recommendations.

| # | Topic Area | # | Recommendation | Recommendation Type |
|---|---|---|---|---|
| 1 | Ensure data alignment in ADS and benchmark data for both crashes and exposure. | 1a. | Reporting differences were considered and addressed, if necessary, through methodological choices. | Required |
| | | 1b. | Exposure differences were considered and addressed, if necessary, through methodological choices. | Recommended |
| | | 1c. | Outcome and exposure units were matched between ADS and benchmark. | Required |
| 2 | Prioritize methodological accuracy but default to conservative analysis. | 2a. | The methodological choices were reasonable and favored conservatism. | Required |
| | | 2b. | Sensitivity analyses, if applicable, were used to explore the effect of methodological decisions on results. | Recommended |
| 3 | Favor outcomes that are directly measurable. | 3a. | Measurable outcomes, if used, were selected with consideration and discussion of potential biases. | Recommended |
| 4 | Consider outcome data transformations to help overcome limitations and address new research questions. | 4a | Transformation methodologies, if used, were well documented, anchored in scientific literature, and appropriate for the evaluation scope. | Required |
| | | 4b | Potential biases and variance introduced by transformation methodologies, if used, were addressed. | Recommended |
| 5 | Quantify uncertainty of the estimates with statistical testing or other methods. | 5a. | Major sources of uncertainty were identified, discussed and/or accounted for in the analysis. | Required |
| | | 5b. | Statistical conclusions, if drawn, were reported following reasonable statistical testing. | Required |

**Table 3.** Transparency specific recommendations.

| # | **Topic Area** | # | **Recommendation** | **Recommendation Type** |
|---|---|---|---|---|
| 6 | Describe the crash and exposure data for both the ADS and the benchmark. | 6a | The names of all data sources relied upon were specified. | Required |
| | | 6b | The origins of the sources were specified. | Required |
| | | 6c | The date ranges of data were specified. | Required |
| | | 6d | The data sampling or reporting frequencies (e.g., annual vs. monthly) were specified. | Required |
| | | 6e | References to any relevant documentation detailing the data source (e.g., data dictionaries) were included. | Required |
| | | 6f | The data sampling scheme (if applicable) was specified. | Recommended |
| | | 6g | Data features influencing inclusion criteria and data reporting were specified. | Recommended |
| | | 6h | Other peculiarities about the data were noted that may have influenced study results. | Recommended |
| 7 | Provide a descriptive summary of the ADS and benchmark data composition. | 7a | Descriptive statistics were presented showing differences in ADS and benchmark sources. | Recommended |
| | | 7b | A table was used to showcase differences between the ADS and benchmark data sources. | Recommended |
| 8 | Describe the ADS deployment being evaluated. | 8a | The ADS systems being evaluated were specified. | Recommended |
| | | 8b | The driving locations (e.g., road type and geographic areas) were specified. | Required |
| | | 8c | Presence of a human operator or supervisor were specified. | Required |
| | | 8d | Other relevant features (if applicable) about the driving environment were specified. | Recommended |
| 9 | Clearly document analysis decisions and steps in the methods section. | 9a | The methodology was described with enough detail to enable replication. | Required |
| | | 9b | Any relied upon published methodology, if applicable, was described and referenced. | Required |
| | | 9c | Any additional data annotations or classifications on top of the raw data were described. | Required |
| 10 | Document limitations in the study's data, methods, scope, and interpretation. | 10a | Potential biases of data source limitations were presented and discussed. | Required |
| | | 10b | Potential biases of analytical decisions were presented and discussed. | Required |
| | | 10c | Limitations around assessment scope were presented and discussed. | Required |
| | | 10d | Effects of the cumulative limitations on results interpretations were presented and discussed. | Recommended |

**Table 4.** Interpretation specific recommendations.

| # | Topic Area | # | Recommendation | Recommendation Type |
|---|---|---|---|---|
| 11 | Build upon past research and justify the scope for what was selected to be studied. | 11a | Relevant literature motivating the study was presented and discussed. | Required |
| | | 11b | Relevant literature related to or influencing the study design choices were presented and discussed. | Required |
| | | 11c | The results and conclusions of the study were presented alongside relevant literature. | Required |
| | | 11d | The justification for excluded outcome measures and analytical lenses were explained. | Recommended |
| 12 | Develop research questions that are clear, concise, and specific. | 12a | Research questions were stated. | Required |
| | | 12b | Research questions were specific and appropriately scoped according to the study design. | Required |
| 13 | Draw appropriate conclusions that accurately reflect the study design. | 13a | The conclusions logically follow the stated research questions. | Required |
| | | 13b | The conclusions stated were restrained to only what could be logically inferred given the study design. | Recommended |
| | | 13c | The conclusions were appropriate given the study's limitations. | Recommended |
| 14 | Exercise caution in relating findings | 14a | Contributing factors were accounted for and discussed in methodology relating findings from lower to higher severity outcomes (if applicable). | Required |
| | between different severity levels. | 14b | Limitations around any methodology relating findings from lower to higher severity outcomes were discussed (if applicable). | Required |
| 15 | Present Rates in Incidents per Exposure Units | 15a. | Rates were presented in terms of incidents per exposure units. | Recommended |

## CONCLUSIONS AND FUTURE CONSIDERATIONS

There is growing research interest in ADS retrospective safety impact assessments. With ongoing research and technological advances in ADS, future contributions to state-of-the art research in this field are anticipated to become exponentially more sophisticated. This publication is not intended to serve as a final statement or minimum or maximum guideline or standard for automated driving systems. Instead, this set of recommendations is proposed as a foundation for designing and evaluating future research in a way that is agnostic to the methodological approach used to overcome the technical challenges that this discipline will face. It is anticipated that future activities, including standards development, will provide further input and improvements. Additionally, stakeholders that are not experts in safety impact assessments may use these recommendations to aid their interpretations of studies they are tasked with evaluating. All stakeholders have a common interest in understanding the safety impact of ADS as an emerging technology. As the adoption of this technology grows, stakeholders require strong evidence on which to base their decisions that have been collected and analyzed using methods to minimize bias.

Because current benchmark data sources were designed under different technological circumstances for a variety of use cases, retrospective safety evaluation of new systems, including ADS, has certain limitations. As a research community, we should build on each other's work, improving collectively. Research findings should serve to extend, validate, and challenge existing knowledge. Gaining consensus, advancing towards standards, and promoting best practices around this work will improve the quality of the research in this area. While not definitive, this publication contributes to the nascent landscape of standardization activities on this topic. In the meantime, these recommendations are expected to generate positive momentum toward high quality, credible research on the safety impact of ADS.

## DISCLOSURE STATEMENT

The positions in this paper do not necessarily reflect the individual positions of any particular contributor and are presented with information known as of the date of publication and are subject to change. No proprietary or confidential information was shared by the parties.